\documentclass[12pt]{article}

\usepackage{amsmath,amsfonts,bm}

\def\eqref#1{equation~\ref{#1}}

\def\1{\bm{1}}

\DeclareMathAlphabet{\mathsfit}{\encodingdefault}{\sfdefault}{m}{sl}
\SetMathAlphabet{\mathsfit}{bold}{\encodingdefault}{\sfdefault}{bx}{n}

\usepackage[left=2.5cm,
right=2.5cm,
top=2.3cm,
bottom=2.3cm,
headheight=20pt,
headsep=10pt,
footskip=25pt,
letterpaper]{geometry}
\usepackage{times}
\usepackage{helvet}
\usepackage{courier}
\usepackage[hyphens]{url}
\usepackage{graphicx}
\usepackage{natbib}
\usepackage{caption}
\usepackage{algorithm}
\usepackage{algorithmic}
\usepackage{booktabs}

\usepackage[most]{tcolorbox}
\usepackage{fancyvrb}
\usepackage{enumitem}
\usepackage{fvextra}
\usepackage{amssymb}
\usepackage{multirow}
\DefineVerbatimEnvironment{Verbatim}{Verbatim}{breaklines=true, breakanywhere=true}

\usepackage{tabularx, array}
\usepackage{ragged2e}
\usepackage{amsmath}
\usepackage{subcaption}
\newcolumntype{Y}{>{\centering\arraybackslash}X}

\graphicspath{{fig/}}

\usepackage{makecell}

\usepackage{newfloat}
\usepackage{listings}
\DeclareCaptionStyle{ruled}{labelfont=normalfont,labelsep=colon,strut=off}
\floatstyle{ruled}
\newfloat{listing}{tb}{lst}{}
\floatname{listing}{Listing}

\usepackage{xcolor}
\usepackage[utf8]{inputenc}
\usepackage[T1]{fontenc}
\usepackage[english]{babel}
\usepackage{amsmath,amsfonts,amssymb,thmtools,amsthm}
\usepackage{hyperref}
\usepackage{fancyhdr}
\usepackage{stfloats}
\usepackage{wrapfig}
\usepackage{epstopdf}
\usepackage[capitalize,noabbrev]{cleveref}
\usepackage{subfloat}
\usepackage{titlesec}
\usepackage{etoolbox}
\usepackage{setspace}
\usepackage{changepage}
\usepackage{svg}
\usepackage[percent]{overpic}
\usepackage[colorinlistoftodos, shadow, color=blue!30!white, disable]{todonotes}
\usepackage{xpatch}
\usepackage{siunitx}

\renewcommand{\sfdefault}{phv}

\renewcommand{\headrulewidth}{0pt}
\renewcommand{\footrulewidth}{0pt}
\fancypagestyle{first}{\fancyfoot[R]{\small\thepage}}

\setlist[itemize]{leftmargin=1em,itemsep=0ex,topsep=0ex}
\titlespacing*{\paragraph}{0pt}{0ex plus .1ex}{1ex}
\titlespacing*{\section}{0ex}{2.3ex plus .3ex minus .0ex}{.6ex plus .3ex minus .2ex}
\titlespacing*{\subsection}{0ex}{1.5ex plus .3ex minus .5ex}{.4ex plus .2ex minus .1ex}
\titlespacing*{\subsubsection}{0ex}{1.2ex plus .3ex minus .3ex}{.3ex plus .2ex minus .2ex}

\xapptocmd\normalsize{%
\abovedisplayskip=.8em plus .2em minus .2em
\belowdisplayskip=.6em plus .1em minus .1em
\abovedisplayshortskip=.8em plus .2em minus .2em
\belowdisplayshortskip=.6em plus .1em minus .1em
}{}{}

\setcitestyle{numbers}
\renewcommand{\cite}[1]{\citep{#1}}

\definecolor{mydarkblue}{rgb}{0.0,0.15,0.7}
\hypersetup{%
colorlinks=true,
linkcolor=mydarkblue,
citecolor=mydarkblue,
filecolor=mydarkblue,
urlcolor=mydarkblue}

\makeatletter

  \renewcommand{\maketitle}{%
    \begingroup
      {\centering\Large\@title\par}%
      \vskip 1em
      \centering
      \begin{tabular}[t]{@{}c@{}}\strut\@author\strut\end{tabular}%
      \vskip 0.3in minus 0.1in
    \endgroup
  }
\makeatother

\newcolumntype{L}{>{\RaggedRight\arraybackslash}X}

\newcommand{\NStanford}{101}
\newcommand{\NFlagged}{149}

\newcommand{\RQSICLRMean}{2.80}

\newcommand{\PctICLRSone}{35}

\newcommand{\RQSICMLMean}{2.83}

\newcommand{\PctICMLSone}{21}

\newcommand{\RQSNeurIPSMean}{2.94}

\newcommand{\PctNeurIPSSone}{20}

\newcommand{\RQSStanfordMean}{3.50}
\newcommand{\PctStanfordStwo}{74}

\newcommand{\RQSHumanMean}{2.86}

\newcommand{\PctHumanStwo}{21}

\newcommand{\PThreeVenues}{0.003}
\newcommand{\PTier}{0.260}
\newcommand{\PLabelVenue}{<10^{-4}}
\newcommand{\PACneg}{0.578}
\newcommand{\CohenDAIvsHuman}{1.26}

\newcommand{\PcaPConeVar}{65}

\newcommand{\PcaMaxCorr}{0.91}

\newcommand{\DimBonfSurvive}{9}
\newcommand{\ACneg}{37}
\newcommand{\ACpeer}{109}
\newcommand{\ACDobs}{-0.03}
\newcommand{\ACMDES}{0.53}

\newcommand{\CIdRqsLo}{+1.19}
\newcommand{\CIdRqsHi}{+1.33}
\newcommand{\DEvid}{+1.57}
\newcommand{\CIEvidLo}{+1.50}
\newcommand{\CIEvidHi}{+1.67}
\newcommand{\DFalsif}{+1.39}
\newcommand{\CIFalsifLo}{+1.31}
\newcommand{\CIFalsifHi}{+1.47}
\newcommand{\DImpr}{-1.23}
\newcommand{\CIImprLo}{-1.38}
\newcommand{\CIImprHi}{-1.08}
\newcommand{\PctReviewsAnyBias}{83}

\newcommand{\TopBiasPhi}{0.14}

\newcommand{\AICoefAdj}{+0.17}
\newcommand{\AICoefAdjLo}{+0.03}
\newcommand{\AICoefAdjHi}{+0.31}
\newcommand{\AICoefAdjP}{0.014}
\newcommand{\LengthCoef}{+0.36}
\newcommand{\NaiveAIGap}{+0.65}
\newcommand{\PctGapExpl}{74}

\newcommand{\ShortThresh}{495}
\newcommand{\NShortReviews}{116}
\newcommand{\RQSShort}{2.66}
\newcommand{\RQSLong}{2.88}
\newcommand{\PctShortSone}{34}
\newcommand{\PctLongSone}{24}

\newcommand{\PctShortTopBias}{59}

\newcommand{\ValN}{76}
\newcommand{\RtAgree}{84}
\newcommand{\RtKappa}{0.77}
\newcommand{\RtAlpha}{0.94}

\newcommand{\XjAgree}{71}
\newcommand{\XjKappa}{0.58}
\newcommand{\XjAlpha}{0.76}

\newcommand{\XvAgree}{33}
\newcommand{\XvKappa}{0.14}
\newcommand{\XvAlpha}{0.23}

\newcommand{\ProbeRealN}{35}
\newcommand{\ProbeRealWin}{33}

\newcommand{\ProbeRealDz}{+2.52}

\newcommand{\ProbeRealSignP}{<.001}

\newcommand{\ProbePctBStwo}{26}

\newcommand{\RQSIclrTwOne}{3.03}
\newcommand{\SoneIclrTwOne}{11}
\newcommand{\StwoIclrTwOne}{23}

\newcommand{\RQSIclrTwTwo}{3.03}

\newcommand{\RQSIclrTwThree}{2.91}

\newcommand{\RQSIclrTwFive}{2.80}
\newcommand{\SoneIclrTwFive}{35}
\newcommand{\StwoIclrTwFive}{16}
\newcommand{\PLongRQS}{<10^{-4}}
\newcommand{\PLongChi}{<10^{-4}}

\newcommand{\LenAdjTwentytwo}{-0.03}
\newcommand{\LenAdjTwentytwoP}{0.20}
\newcommand{\LenAdjTwentythree}{-0.15}
\newcommand{\LenAdjTwentythreeP}{<10^{-8}}
\newcommand{\LenAdjTwentyfive}{-0.21}

\newcommand{\LenAdjTwentyfiveP}{<10^{-11}}

\title{Articulate Intuition or Genuine Analysis? 
Benchmarking \\ Epistemic Reliability in LLM-as-a-Judge Peer Review}

\date{}
\author{
Nuo Chen \quad
Qian Wang \quad
Qingyun Zou \quad
Bingsheng He \vspace{10pt} \\
National University of Singapore
}

\begin{document}
\pagestyle{fancy}

\maketitle
\thispagestyle{first}


\begin{abstract}
When an LLM judge calls a peer review ``analytical'' and a human committee calls another review ``high quality,'' are they tracking the same thing? We argue they are not, and that the difference matters philosophically. We operationalise Kahneman's dual-process theory into a structured rubric for peer review and release \textit{Kahneman4Review}, a benchmark of $3{,}563$ rated reviews scored along nine theoretically motivated textual dimensions, eight bias diagnostics, and a continuous reasoning-quality score. Three findings bear on \emph{trustworthiness}: decision tier is not detectably aligned with the rubric's text-grounded epistemic-quality proxy; public-showcase agentic reviews receive higher raw scores than pooled human reviews, but length and venue explain most of the gap and the samples are not paper-paired; and ICLR review-text diagnostics shift at the 2022--2023 transition, temporally coincident with widespread LLM availability but without identifying its cause. A matched function-probe pilot further shows that the rubric distinguishes textual probes designed to contrast genuine fault-finding with surface fluency. We argue that a trustworthy reliability benchmark for LLM judges must separate \emph{analytical form} from \emph{epistemic function}, and propose concrete design choices toward that goal. An interactive demo is available at \href{https://huggingface.co/spaces/nuojohnchen/Kahneman4Review}{https://huggingface.co/spaces/nuojohnchen/Kahneman4Review}.
\end{abstract}

\section{Introduction}
\label{sec:intro}

Peer review is the most widely used epistemic gatekeeping mechanism in science, and large language models are increasingly proposed both to \emph{produce} reviews \cite{liang2023can,paperreviewai} and to \emph{judge} review quality \cite{zheng2023judging,chiang2023cangpt}. Deploying LLMs in either role invites a philosophical question that the machine learning community rarely poses directly: \textit{what counts as a good review?} An answer is usually smuggled in through training objectives and benchmark design (acceptance prediction, score agreement, human-preference alignment), yet philosophers have long argued that epistemic quality is not a scalar outcome but a property of the reasoning itself \cite{hempel1948,lipton2004,elgin2017}. A review may be technically correct and still textually under-justified; a negative review may be high-quality if it is grounded, falsifiable, and responsive to rebuttal.

Peer review offers an unusually clean test bed for this problem. First, the review is a written artefact amenable to fine-grained analysis. Second, there exist outcomes (acceptance, meta-review, AC flags) that are typically treated as ground truth but which our data will show are \emph{not detectably aligned} with the rubric's text-grounded epistemic-quality proxy. Third, the recent appearance of agentic AI reviewers \cite{paperreviewai,liang2023can} creates a natural textual contrast between reviews written by humans under deadline and outputs from retrieval-augmented LLM pipelines with full paper access. The contrast motivates a dual-process-inspired question for ML evaluation: can an LLM judge identify textual evidence that separates \emph{articulate intuition} from \emph{genuine analysis} without mistaking analytical form for verified epistemic function?

\textbf{Contributions.}
\textbf{(1)}~We propose nine theoretically motivated textual dimensions, eight bias diagnostics, a four-label taxonomy, and a continuous reasoning-quality score, grounded in analytic epistemology and philosophy of explanation.
\textbf{(2)}~We rate $3{,}563$ reviews using \texttt{claude-sonnet-4-6} as the judge: $1{,}155$ stratified-sample human reviews from ICLR, ICML, and NeurIPS 2025; $2{,}307$ ICLR 2021/2022/2023 human reviews for the longitudinal study; and $\NStanford{}$ reviews from an open agentic AI-reviewer pipeline. The released code, rubric prompt, ratings, and provenance metadata are available through the \href{https://huggingface.co/spaces/nuojohnchen/Kahneman4Review}{Kahneman4Review demo}.
\textbf{(3)}~We report three empirical findings bearing directly on \textit{trustworthiness}: (a) analytical-trace scores exhibit a field-wide ceiling across venues (RQS $\in [\RQSICLRMean, \RQSNeurIPSMean]$); (b) decision tier does not predict RQS; (c) public-showcase agentic and pooled human reviews are statistically separable on precisely those dimensions trained LLMs can fluently imitate.
\textbf{(4)}~Although the rubric proposes nine theoretically distinct dimensions, the current LLM judge collapses them into one dominant analytical-trace factor. We treat this as a \emph{failure-mode finding} for textual benchmarks, not evidence of multidimensional measurement.

\section{Related Work}
\label{sec:related}

\textbf{Peer-review analysis} has focused largely on outcome signals: score agreement, bias by author identity, and strategic behaviour \cite{tomkins2017reviewer,stelmakh2021,shah2022peer,langford2015}. These framings implicitly equate review reliability with inter-rater consistency or score-validity. We instead operationalise review quality as a property of the reasoning text.

\textbf{LLM-as-a-judge} systems are used for preference scoring \cite{zheng2023judging,chiang2023cangpt,liu2024calibrating}, and recent work documents their systematic biases, including self-recognition and in-distribution favoritism \cite{panickssery2024llm}. Concurrent work \cite{liang2023can} uses an LLM to \emph{generate} review feedback; our rubric is instead a structured \emph{judge} whose outputs are auditable per dimension and per span, so the question of judge reliability can be posed at finer granularity than scalar agreement.

\textbf{Philosophy and ML evaluation.} Work in interpretable ML has increasingly borrowed from philosophy and cognitive science \cite{miller2019,doshivelez2017}. Our framing draws on analytic accounts of explanation \cite{hempel1948,lipton2004}, understanding \cite{elgin2017}, and epistemic luck \cite{pritchard2005}, and connects to FAccT-adjacent discussions of trust and reliability \cite{jacovi2021trust} that treat trustworthiness as more than predictive accuracy.

\section{A Dual-Process Rubric for Review Epistemics}
\label{sec:method}

\textbf{From theory to rubric.} \citet{kahneman2011} distinguishes \textit{System 1} (fast, associative, impression-based) from \textit{System 2} (slow, deliberative, evidence-integrating). We use this distinction as inspiration for textual diagnostics of epistemic reasoning: System-1-like traces include compressed impressions and unsupported labels, whereas System-2-like traces include explicit evidence linkage, rebuttability, and uncertainty handling \cite{evans2013,gigerenzer2011}. These labels describe textual traces relevant to articulate intuition and genuine analysis, not the reviewer's cognitive state. Without paper access, they cannot establish that genuine analysis occurred or that a critique is correct.

\textbf{Dimensions.} Each review is scored on nine 0--3 dimensions. Two (\textit{impression reliance, heuristic shortcut usage}) are System-1-valenced; the other seven are System-2-valenced. Each dimension anchors to a specific philosophical desideratum: \textit{falsifiability} to \citet{popper1959}, \textit{evidence specificity} to inference-to-the-best-explanation \cite{lipton2004}, \textit{uncertainty handling} to the contrastive view of understanding \cite{elgin2017}, \textit{issue prioritisation} to virtue epistemology \cite{sosa2007}, \textit{reasoning density} to empirical adequacy \cite{vanfraassen1980}, \textit{reasoning chain} to the structure of explanation \cite{hempel1948}, \textit{heuristic shortcut} to fast-and-frugal cognition \cite{gigerenzer2011}. The full map, with rubric cues, is in \cref{tab:dim-phil}.

\begin{table}[!ht]
\caption{Nine rubric dimensions and their philosophical anchor points. The third column is the rubric cue the judge looks for.}
\label{tab:dim-phil}
\begin{center}\begin{small}
\begin{tabular}{@{}lll@{}}
\toprule
Dimension & Anchor & Cue \\
\midrule
Impression rel.$^\dagger$     & intuitive judgment \cite{kahneman2011}      & global labels (``novel'') \\
Reasoning chain               & argument structure \cite{hempel1948}        & premise $\to$ conclusion \\
Evidence specificity          & IBE \cite{lipton2004}                       & numbers, eqns., quotes \\
Falsifiability                & demarcation \cite{popper1959}               & concrete refutability \\
Heuristic use$^\dagger$       & fast-and-frugal \cite{gigerenzer2011}       & venue-fit shortcuts \\
Uncertainty handling          & understanding \cite{elgin2017}              & conditionals, updating \\
Issue prioritisation          & virtue epist.\ \cite{sosa2007}              & core vs.\ peripheral \\
Reasoning density             & empirical adequacy \cite{vanfraassen1980}   & support per claim \\
Core criticality              & rebuttability \cite{popper1959,lipton2004}  & decision-moving \\
\bottomrule
\end{tabular}
\end{small}\end{center}
{\footnotesize $^\dagger$System-1-valenced: low scores indicate analytical reasoning.}
\end{table}

\textbf{Continuous scores.} The judge emits two textual-trace scalars $s_1,s_2\!\in\![1,5]$ and a Reasoning Quality Score $\text{RQS}\!\in\![1,5]$ intended to separate impression-like form from explicitly justified form. The label $\ell\in\{\textsc{System 1}, \textsc{System 2}, \textsc{Mixed}, \textsc{Non-evaluative}\}$ is a discretisation of the $(s_1, s_2, \text{RQS})$ joint. We explicitly require a \emph{nearest alternative label} and a counterfactual justification (``why not X?''), which is the rubric's nod to contrastive explanation \cite{lipton2004,miller2019}.

\textbf{Bias diagnostics.} Eight bias categories (\textit{Checklist Inflation, Representativeness, Question Substitution, Conclusion-First, Overconfidence, Narrative Fallacy, Authority Substitution, Confirmation Bias}) are detected as open-set tags.

\textbf{What does the rubric actually measure?} A principal-components analysis of the nine dimensions across the $1{,}155$ cross-venue human ratings finds a single dominant factor explaining \PcaPConeVar\% of variance; one eigenvalue exceeds the Kaiser threshold, maximum off-diagonal correlation is $\PcaMaxCorr$, and PC1 loads positively on every sign-flipped dimension (\cref{fig:pca}). Thus, although the rubric is theoretically multidimensional, the present judge does not recover nine empirically distinct constructs. We treat this as a \emph{failure-mode finding}: the ratings collapse to a near-unidimensional analytical-trace axis, so a well-prompted LLM may saturate the benchmark globally rather than on one axis (returned to in \cref{sec:discussion}).

\section{Empirical Study}
\label{sec:exp}

\subsection{Data and protocol}
\label{sec:data-protocol}

We stratified-sample 100 papers per venue from ICLR, ICML, NeurIPS 2025 by tier (6 oral, 10 spotlight, 84 poster), yielding $1{,}155$ human reviews; we additionally rate 200 random papers each from ICLR 2021/2022/2023 ($2{,}307$ further reviews) under the same judge for the longitudinal study, and $\NStanford{}$ reviews from a public showcase of an open agentic-reviewer pipeline \cite{paperreviewai,fars2026} ($3{,}563$ ratings in total) (full provenance and the curation caveat are in \cref{app:data}). All ratings use \texttt{claude-sonnet-4-6} with a frozen prompt. The LLM judge rates each review from its text, venue, and rating metadata alone; the reviewed paper is not included in the judge's input. We use Mann--Whitney U, Kruskal--Wallis, one-way ANOVA, and $\chi^2$ as appropriate; per-dimension comparisons use Bonferroni correction for 9 tests.

\begin{figure}[!ht]
\centering
\includegraphics[width=0.6\linewidth]{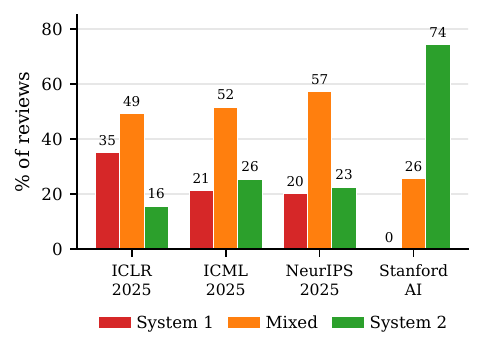}
\caption{Label distribution across three human venues and the Stanford agentic reviewer. Venue-level differences are significant ($\chi^2,\ p\PLabelVenue$).}
\label{fig:label-dist}
\end{figure}

\subsection{Analytical-trace scores show a field-wide ceiling and temporal decline}
\label{sec:ceiling}

Mean RQS in 2025 is $\RQSICLRMean{}$ (ICLR), $\RQSICMLMean{}$ (ICML), $\RQSNeurIPSMean{}$ (NeurIPS), on a 1--5 scale; venue-level Kruskal--Wallis is significant ($p=\PThreeVenues$) but the absolute spread is small. ICLR is a statistical outlier in \textit{label} distribution: $\PctICLRSone\%$ of ICLR reviews are System~1 vs.\ $\PctICMLSone\%$ (ICML) and $\PctNeurIPSSone\%$ (NeurIPS) (\cref{fig:label-dist}).

\begin{figure}[!ht]
\centering
\includegraphics[width=0.92\linewidth]{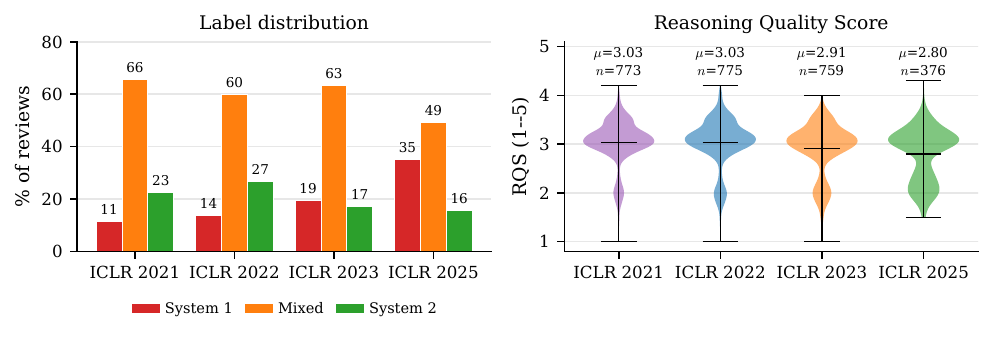}
\caption{ICLR review-text diagnostics across years. Left: label distribution shifts from fewer System-1-like and more System-2-like traces (2021/2022) toward a more System-1-like mix (2025). Right: RQS distribution declines monotonically. K--W $p\PLongRQS$; length-adjusted residual gap $\LenAdjTwentyfive$ RQS, $p\LenAdjTwentyfiveP$.}
\label{fig:longitudinal}
\end{figure}

\textbf{Temporal trend.} Rating ICLR 2021, 2022, and 2023 with the same judge and prompt as 2025 (total $n{=}2{,}683$ reviews) reveals a discontinuity at the 2022--2023 transition, temporally coincident with widespread LLM availability: mean RQS is identical in 2021 and 2022 ($\RQSIclrTwOne$ vs $\RQSIclrTwTwo$), then drops to $\RQSIclrTwThree$ (2023) and $\RQSIclrTwFive$ (2025); the System-1-like share triples, $\SoneIclrTwOne\% \to \SoneIclrTwFive\%$; the System-2-like share nearly halves, $\StwoIclrTwOne\% \to \StwoIclrTwFive\%$ (\cref{fig:longitudinal}). K--W $p\PLongRQS$, label $\chi^2$ $p\PLongChi$. Length-controlled regression with 2021 as reference and $\log$-length included puts \emph{2022 vs 2021 at $\LenAdjTwentytwo$ RQS ($p{=}\LenAdjTwentytwoP$, n.s.)} but \emph{2023 vs 2021 at $\LenAdjTwentythree$ ($p\LenAdjTwentythreeP$)} and \emph{2025 vs 2021 at $\LenAdjTwentyfive$ ($p\LenAdjTwentyfiveP$)}. This pattern is descriptive, not causal: reviewer composition, review policy, conference scale, changing norms, and LLM-assisted drafting are all possible explanations.

\subsection{Outcomes do not predict the epistemic-quality proxy}
\label{sec:tier-null}

\begin{figure}[!h]
\centering
\includegraphics[width=0.6\linewidth]{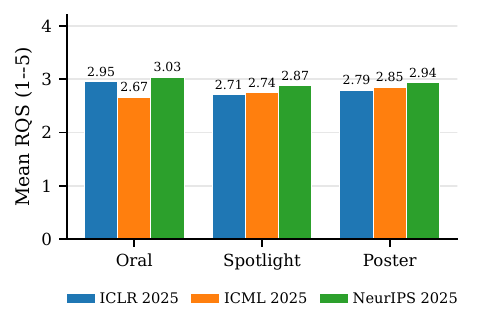}
\caption{Mean RQS by acceptance tier per venue. Sample sizes: ICLR/ICML/NeurIPS Oral $n=23$/$23$/$26$, Spotlight $36$/$40$/$43$, Poster $317$/$313$/$334$. Within-venue differences are within standard error; pooled one-way ANOVA is null ($p=\PTier$).}
\label{fig:rqs-tier}
\end{figure}

We find no detectable main effect of acceptance tier on RQS, pooled across venues ($p=\PTier$, \cref{fig:rqs-tier}); within each venue, tier means fall within 0.2 RQS points. This null is informative under either reading: \emph{either} the conference system's most visible outcome signal does not track the rubric's textual proxy, \emph{or} the rubric and outcome are tracking related properties at a level our $n$ cannot resolve. In either case, tier agreement is at best a weak proxy for articulated reasoning quality in our data.

\begin{figure}[!ht]
\centering
\includegraphics[width=0.6\linewidth]{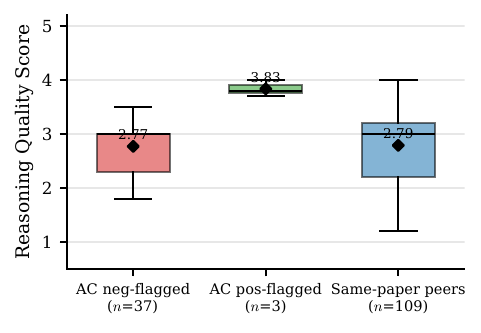}
\caption{RQS distribution by AC annotation group from in-metareview commentary. The null result ($p=\PACneg{}$, observed $d=\ACDobs{}$ against MDES $\ACMDES{}$) shows AC scalar judgments and rubric-level signals are not strongly aligned in our data; the sample is underpowered to distinguish true orthogonality from a small effect.}
\label{fig:ac-validation}
\end{figure}

As an auxiliary probe, we rate \NFlagged{} reviewers whose quality was explicitly commented on by area chairs. AC-flagged negatively-commented reviewers ($n{=}\ACneg$) and their same-paper peers ($n{=}\ACpeer$) differ on RQS at $p=\PACneg$ (n.s.), observed $d=\ACDobs$ against a minimum-detectable effect of $d=\ACMDES$ at 80\% power. We do not read this as proof of orthogonality (our sample is underpowered for $|d|<0.5$), only as evidence that AC-level scalar judgments and rubric-level signals are not strongly aligned in our data (\cref{fig:ac-validation}).

\subsection{Agentic AI reviews exhibit more System-2-like traces: mostly via length}
\label{sec:ai-vs-human}

\begin{figure}[!ht]
\centering
\begin{subfigure}[t]{0.30\linewidth}
\centering
\includegraphics[width=\linewidth]{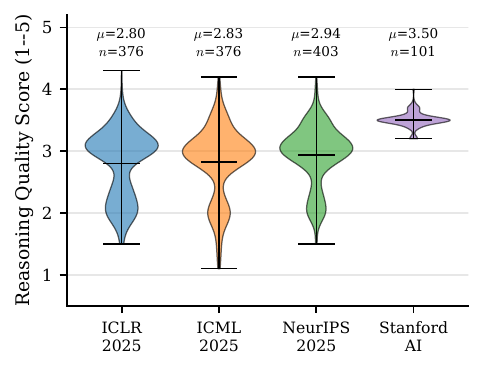}
\caption{}
\label{fig:rqs-violin}
\end{subfigure}
\hfill
\begin{subfigure}[t]{0.68\linewidth}
\centering
\includegraphics[width=\linewidth]{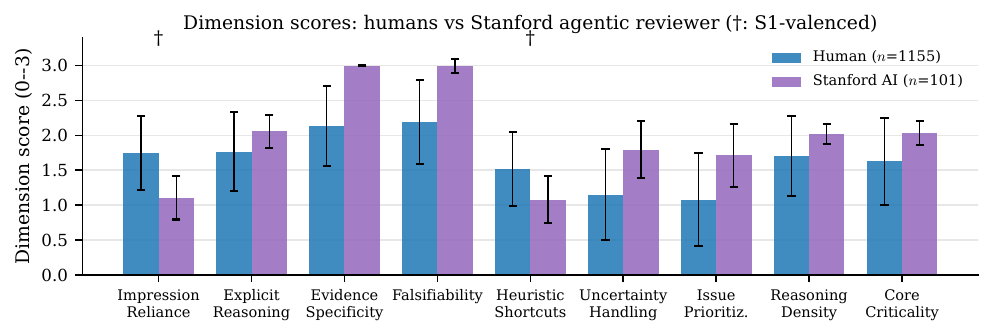}
\caption{}
\label{fig:dim-compare}
\end{subfigure}
\caption{Human reviewers vs.\ the Stanford agentic reviewer. \textbf{(a)} RQS distribution per group (violin, mean line): the public-showcase agentic distribution has higher raw RQS (mean 3.50, SD 0.12), while humans are broader (means $\in[2.80, 2.94]$, SDs $\in[0.48, 0.57]$). This unpaired descriptive comparison does not establish reviewer competence. \textbf{(b)} Mean score per rubric dimension; error bars $\pm 1$ SD; ${\dagger}$ marks System-1-valenced dimensions, for which lower is more analytical. All 9 dimensions differ at $p<0.05$ after Bonferroni correction, with $|d|\ge 0.54$ in every case.}
\label{fig:ai-vs-human}
\end{figure}

Stanford agentic-reviewer outputs are classified System~2 at $\PctStanfordStwo\%$ versus $\PctHumanStwo\%$ for pooled humans. Mean RQS is $\RQSStanfordMean$ versus $\RQSHumanMean$: Cohen's $d=\CohenDAIvsHuman$ with 95\% bootstrap CI $[\CIdRqsLo, \CIdRqsHi]$ (2000 resamples). Per-dimension effect sizes are large and tightly bounded: \emph{evidence specificity} $d=\DEvid$ [\CIEvidLo, \CIEvidHi], \emph{falsifiability} $d=\DFalsif$ [\CIFalsifLo, \CIFalsifHi], and \emph{impression reliance} $d=\DImpr$ [\CIImprLo, \CIImprHi] (\cref{fig:dim-compare}). \textit{All \DimBonfSurvive/9 dimensions remain significant at $\alpha{=}0.05$ after Bonferroni correction.} Across the full RQS distributions (\cref{fig:rqs-violin}), the agentic reviewer's scores are tight and elevated, while the human venues are broader and centred lower.

Because these public-showcase outputs are potentially curated and not paper-paired, this result is not evidence that AI reviewers are epistemically better than human reviewers.

\textit{Length is a confound.} Agentic outputs are systematically longer than human reviews. To separate the analytical-trace score from verbosity we fit
\begin{equation}
\text{RQS}_i = \beta_0 + \beta_1\,\mathbb{1}[\text{AI}]_i + \beta_2 \log(\text{length}_i) + \boldsymbol{\gamma}^\top \boldsymbol{V}_i + \varepsilon_i,
\label{eq:lengthreg}
\end{equation}
where $\boldsymbol{V}_i$ is a one-hot venue vector (Stanford = reference). This shrinks the AI coefficient from a naive $\NaiveAIGap$ to $\hat\beta_1 = \AICoefAdj$ (95\% CI $[\AICoefAdjLo, \AICoefAdjHi]$, $p{=}\AICoefAdjP$), with $\hat\beta_2 = \LengthCoef$ per log-character; length and venue jointly explain $\PctGapExpl\%$ of the naive gap. The residual signal is real but small, and the ``System 2'' label should be read with this control in mind.

\begin{figure}[!ht]
\centering
\includegraphics[width=0.7\linewidth]{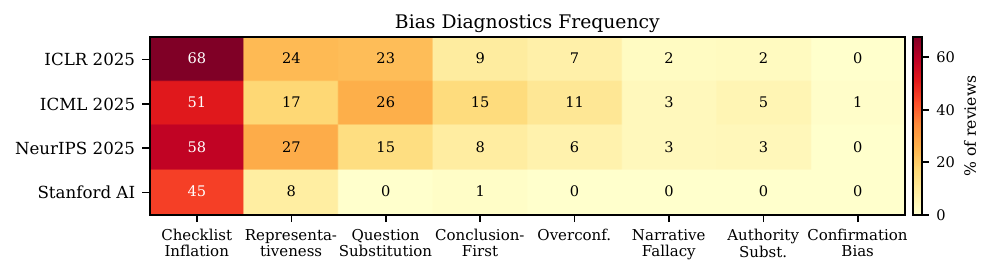}
\caption{Bias-diagnostic frequency (\% of reviews in which the tag appears), per group. \textit{Checklist Inflation} dominates in all human venues; the agentic reviewer is also dominated by \textit{Checklist Inflation} (45\%, less than humans) with some \textit{Representativeness Heuristic} (8\%) and essentially no \textit{Question Substitution} or \textit{Authority Substitution}.}
\label{fig:bias-heatmap}
\end{figure}

\textit{Bias patterns.} Bias tags co-occur non-trivially: \PctReviewsAnyBias\% of human reviews carry at least one bias tag, and the strongest pairwise association ($\phi=\TopBiasPhi$) is between Representativeness Heuristic and Question Substitution (full $\phi$ matrix in \cref{fig:bias-cooccur}). This is a modest co-occurring System-1-like trace pattern: a review that substitutes an easier question for the one posed by the paper also tends to answer it via representativeness. The Stanford agentic reviewer is dominated by Checklist Inflation (45\%) with some Representativeness (8\%) and never Question Substitution or Authority Substitution (\cref{fig:bias-heatmap}).

This AI-vs-human pattern, with the sampling and length-control caveats, is the paper's central diagnostic finding.

\textit{Measurement validation.} On a fixed $n{=}100$ subset (\cref{app:validation} for the protocol), frozen-prompt test-retest of \texttt{claude-3-7-sonnet} gives $\kappa{=}\RtKappa$, $\alpha_{\text{ord}}{=}\RtAlpha$; cross-judge vs.\ \texttt{gpt-4o} gives $\kappa{=}\XjKappa$, $\alpha{=}\XjAlpha$ (\cref{tab:validation}). The earlier $\kappa{=}0.09$ figure was a prompt-revision artefact. Cross-\emph{version} agreement is the weakest link ($\kappa{=}\XvKappa$ between \texttt{sonnet-4-6} and \texttt{sonnet-3.7}), so reliability claims about LLM judges should be reported as model-conditional throughout.

\begin{table}[!ht]
\centering
\caption{Inter-run, cross-judge, and cross-version agreement on the fixed $n{=}100$ validation subset, with the original \texttt{claude-sonnet-4-6} ratings as a comparator. After dropping malformed-JSON outputs, $n{=}\ValN$ reviews are paired across all three rating conditions.}
\label{tab:validation}
\begin{tabular}{@{}lcccc@{}}
\toprule
Comparison & $n$ & label agree & $\kappa$ & $\alpha_{\text{ord}}$(RQS) \\
\midrule
Sonnet-3.7 run1 vs.\ run2 (test-retest) & \ValN & \RtAgree\% & \RtKappa & \RtAlpha \\
Sonnet-3.7 vs.\ GPT-4o (cross-judge)    & 84    & \XjAgree\% & \XjKappa & \XjAlpha \\
Sonnet-3.7 vs.\ Sonnet-4-6 (cross-version) & 85 & \XvAgree\% & \XvKappa & \XvAlpha \\
\bottomrule
\end{tabular}
\end{table}

\textit{Function-probe pilot.} For $n{=}\ProbeRealN$ ICLR-2025 papers, \texttt{claude-3-7-sonnet} identifies one falsifiable concern and emits two $\sim$250-word reviews matched on length and specificity markers: review~A's primary weakness IS that concern, B's are surface peripheral asks. Result: \textbf{$\ProbeRealWin/\ProbeRealN$ have $\text{RQS}(A){>}\text{RQS}(B)$}, $d_z{=}\ProbeRealDz$, sign-test $p\ProbeRealSignP$; all A-reviews labeled \textsc{System 2}, only \ProbePctBStwo\% of B's. Caveat: same model family generates and rates; curated public-errata faults are the necessary follow-up.

\section{Discussion: Analytical Form and Epistemic Function}
\label{sec:discussion}

The rubric separates venues and yields auditable dimension- and span-level outputs, but it also gives public-showcase agentic outputs high scores on dimensions a well-prompted LLM can manufacture. Because PCA shows these dimensions moving together, we treat the rubric as a \emph{partially-validated diagnostic of textual-benchmark failure}, not a settled reliability metric.

An LLM reviewer that fluently produces the \textit{form} of analytical discourse without establishing that genuine analysis was performed is the modern epistemic analogue of an empirically adequate theory that is not explanatory \cite{hempel1948,vanfraassen1980,lipton2004}; \citet{elgin2017}'s account of understanding as competence-over-a-domain says the relevant competence is saying what would change one's verdict: \emph{attempting to falsify}, not paraphrase. Our probe pilot shows that the rubric distinguishes textual signatures designed to contrast genuine fault-finding with surface fluency under matched form, but the AI-vs-human gap indicates that much of the headline signal still rides on analytical form rather than verified epistemic function.

\textbf{What a trustworthy LLM judge would look like.} Three design choices suggested by the results: (i) \emph{Audit by span, not by score}, since judgments without text spans are unfalsifiable \cite{popper1959}; (ii) \emph{Require a counterfactual label}, forcing contrastive reasoning \cite{lipton2004,miller2019}; (iii) \emph{Separate analytical form from epistemic function via probes}: our $n{=}\ProbeRealN$ pilot finds the rubric discriminates $\ProbeRealWin/\ProbeRealN$ matched pairs in favour of the genuine-fault condition ($d_z{=}\ProbeRealDz$, $p\ProbeRealSignP$), and scaling to a curated 200-item probe set with public-errata faults is the natural next step.

\section{Conclusion}
\label{sec:conclusion}

Frozen-prompt ratings are stable and discriminate $\ProbeRealWin/\ProbeRealN$ matched-form probes, but decision tier does not predict RQS and length and venue explain $\PctGapExpl\%$ of the naive AI--human gap. These scores capture articulated traces, not verified epistemic function: without the reviewed paper, they cannot establish genuine analysis, critique correctness or fidelity, or reviewer competence. Further limitations are judge self-favoritism \cite{panickssery2024llm}, model-version sensitivity ($\kappa{=}\XvKappa$), and omitted social epistemology and causal modelling; future work needs model-conditional reporting, public-errata probes, and expert anchors.

\section*{Impact Statement}
This work aims to make LLM-assisted peer review more auditable. A judge rewarding only linguistic surface form could legitimise AI review pipelines whose analytical competence is overstated; function probes and span-level audits are intended to resist that risk.

\clearpage
\begin{hyphenrules}{nohyphenation}
\setlength{\bibsep}{.5ex plus .8ex}
\bibliographystyle{unsrtnat}
\nocite{langford2015}
\bibliography{main}

@book{kahneman2011,
  author    = {Kahneman, Daniel},
  title     = {Thinking, Fast and Slow},
  publisher = {Farrar, Straus and Giroux},
  year      = {2011},
  address   = {New York}
}

@book{evans2013,
  author    = {Evans, Jonathan St.\ B.\ T. and Stanovich, Keith E.},
  title     = {Dual-Process Theories of Higher Cognition: Advancing the Debate},
  publisher = {Perspectives on Psychological Science},
  year      = {2013}
}

@book{elgin2017,
  author    = {Elgin, Catherine Z.},
  title     = {True Enough},
  publisher = {MIT Press},
  year      = {2017}
}

@book{pritchard2005,
  author    = {Pritchard, Duncan},
  title     = {Epistemic Luck},
  publisher = {Oxford University Press},
  year      = {2005}
}

@book{sosa2007,
  author    = {Sosa, Ernest},
  title     = {A Virtue Epistemology: Apt Belief and Reflective Knowledge},
  publisher = {Oxford University Press},
  year      = {2007}
}

@article{hempel1948,
  author  = {Hempel, Carl G. and Oppenheim, Paul},
  title   = {Studies in the Logic of Explanation},
  journal = {Philosophy of Science},
  volume  = {15},
  number  = {2},
  pages   = {135--175},
  year    = {1948}
}

@book{lipton2004,
  author    = {Lipton, Peter},
  title     = {Inference to the Best Explanation},
  publisher = {Routledge},
  edition   = {2nd},
  year      = {2004}
}

@article{miller2019,
  author  = {Miller, Tim},
  title   = {Explanation in Artificial Intelligence: Insights from the Social Sciences},
  journal = {Artificial Intelligence},
  volume  = {267},
  pages   = {1--38},
  year    = {2019}
}

@article{doshivelez2017,
  author  = {Doshi-Velez, Finale and Kim, Been},
  title   = {Towards a Rigorous Science of Interpretable Machine Learning},
  journal = {arXiv preprint arXiv:1702.08608},
  year    = {2017}
}

@inproceedings{zheng2023judging,
  author    = {Zheng, Lianmin and Chiang, Wei-Lin and Sheng, Ying and Zhuang, Siyuan and Wu, Zhanghao and Zhuang, Yonghao and Lin, Zi and Li, Zhuohan and Li, Dacheng and Xing, Eric P. and Zhang, Hao and Gonzalez, Joseph E. and Stoica, Ion},
  title     = {Judging LLM-as-a-Judge with MT-Bench and Chatbot Arena},
  booktitle = {Advances in Neural Information Processing Systems (NeurIPS)},
  year      = {2023}
}

@inproceedings{liu2024calibrating,
  author    = {Liu, Yinhong and Zhou, Han and Guo, Zhijiang and Shareghi, Ehsan and Vuli\'{c}, Ivan and Korhonen, Anna and Collier, Nigel},
  title     = {Aligning with Human Judgement: The Role of Pairwise Preference in Large Language Model Evaluators},
  booktitle = {Conference on Language Modeling (COLM)},
  year      = {2024}
}

@inproceedings{stelmakh2021,
  author    = {Stelmakh, Ivan and Shah, Nihar B. and Singh, Aarti},
  title     = {Catch Me if {I} Can: Detecting Strategic Behaviour in Peer Assessment},
  booktitle = {Proceedings of the AAAI Conference on Artificial Intelligence},
  year      = {2021}
}

@article{shah2022peer,
  author  = {Shah, Nihar B.},
  title   = {Challenges, Experiments, and Computational Solutions in Peer Review},
  journal = {Communications of the ACM},
  volume  = {65},
  number  = {6},
  pages   = {76--87},
  year    = {2022},
  doi     = {10.1145/3528086}
}

@article{langford2015,
  author  = {Langford, John and Guzdial, Mark},
  title   = {The Arbitrariness of Reviews, and Advice for School Administrators},
  journal = {Communications of the ACM},
  volume  = {58},
  number  = {4},
  pages   = {12--13},
  year    = {2015},
  doi     = {10.1145/2732417}
}

@article{tomkins2017reviewer,
  author  = {Tomkins, Andrew and Zhang, Min and Heavlin, William D.},
  title   = {Reviewer Bias in Single- versus Double-blind Peer Review},
  journal = {Proceedings of the National Academy of Sciences (PNAS)},
  volume  = {114},
  number  = {48},
  pages   = {12708--12713},
  year    = {2017}
}

@misc{paperreviewai,
  author = {Jiang, Yixing and Ng, Andrew},
  title  = {Stanford Agentic Reviewer},
  howpublished = {\url{https://paperreview.ai}},
  year   = {2025},
  note   = {Stanford Machine Learning Group. Accessed 2026-06-12}
}

@misc{fars2026,
  author = {{Analemma Intelligence}},
  title  = {{FARS}: Fully Automated Research System},
  howpublished = {\url{https://analemma.ai/fars/}},
  year   = {2026},
  note   = {Public showcase corpus; reviews generated by the Stanford Agentic Reviewer (paperreview.ai). Accessed 2026-06-12}
}

@inproceedings{chiang2023cangpt,
  author    = {Chiang, Cheng-Han and Lee, Hung-Yi},
  title     = {Can Large Language Models Be an Alternative to Human Evaluations?},
  booktitle = {Annual Meeting of the Association for Computational Linguistics (ACL)},
  year      = {2023}
}

@article{liang2023can,
  author  = {Liang, Weixin and Zhang, Yuhui and Cao, Hancheng and Wang, Binglu and Ding, Daisy Y. and Yang, Xinyu and Vodrahalli, Kailas and He, Siyu and Smith, Daniel S. and Yin, Yian and McFarland, Daniel A. and Zou, James},
  title   = {Can Large Language Models Provide Useful Feedback on Research Papers? A Large-Scale Empirical Analysis},
  journal = {NEJM AI},
  volume  = {1},
  number  = {8},
  year    = {2024},
  doi     = {10.1056/AIoa2400196}
}

@article{gigerenzer2011,
  author  = {Gigerenzer, Gerd and Gaissmaier, Wolfgang},
  title   = {Heuristic Decision Making},
  journal = {Annual Review of Psychology},
  volume  = {62},
  pages   = {451--482},
  year    = {2011}
}

@book{popper1959,
  title     = {The Logic of Scientific Discovery},
  author    = {Popper, Karl},
  year      = {1959},
  publisher = {Hutchinson},
  address   = {London}
}

@book{vanfraassen1980,
    author = {Fraassen, Bas. C. van},
    title = {The Scientific Image},
    publisher = {Oxford University Press},
    year = {1980},
    month = {12},
    isbn = {9780198244271},
    doi = {10.1093/0198244274.001.0001},
    url = {https://doi.org/10.1093/0198244274.001.0001},
}

@inproceedings{jacovi2021trust,
  author    = {Jacovi, Alon and Marasovi\'{c}, Ana and Miller, Tim and Goldberg, Yoav},
  title     = {Formalizing Trust in Artificial Intelligence: Prerequisites, Causes and Goals of Human Trust in {AI}},
  booktitle = {Proceedings of the 2021 ACM Conference on Fairness, Accountability, and Transparency (FAccT)},
  year      = {2021},
  doi       = {10.1145/3442188.3445923}
}

@inproceedings{panickssery2024llm,
  author    = {Panickssery, Arjun and Bowman, Samuel R. and Feng, Shi},
  title     = {{LLM} Evaluators Recognize and Favor Their Own Generations},
  booktitle = {Advances in Neural Information Processing Systems (NeurIPS)},
  year      = {2024}
}
\end{hyphenrules}

\clearpage
\appendix

\section{Rubric details}
\label{app:rubric}

\textbf{Labels.} \textsc{System 1}, \textsc{System 2}, \textsc{Mixed}, \textsc{Non-evaluative}. A review is labelled \textsc{System 1} if its text is dominated by impression-like, compressed, and conclusion-first traces; \textsc{System 2} if it is dominated by explicit, evidence-grounded, and uncertainty-sensitive traces; \textsc{Mixed} if both textual patterns are substantively present; and \textsc{Non-evaluative} if it is administrative or clarificatory rather than judgemental. These labels do not identify the reviewer's cognitive state.

\textbf{Dimensions (0--3).}
\begin{itemize}
\item \textit{Impression reliance} (S1-valenced): reliance on global labels and unsupported overall claims.
\item \textit{Explicit reasoning chain}: observable premise--conclusion structure.
\item \textit{Evidence specificity}: citation of concrete figures, numbers, equations, or prior work passages.
\item \textit{Falsifiability/rebuttability}: whether each concern admits a concrete refutation.
\item \textit{Heuristic shortcut usage} (S1-valenced): venue-fit, familiarity, authority shortcuts.
\item \textit{Uncertainty handling \& belief updating}: hedging, conditioning, responsiveness to rebuttal in principle.
\item \textit{Issue prioritisation}: acceptance-critical vs.\ peripheral weighting.
\item \textit{Reasoning density}: analytical support per sentence.
\item \textit{Core criticality}: whether weaknesses would actually move a decision if true.
\end{itemize}

\textbf{Continuous scalars.} $s_1, s_2 \in [1,5]$ are independent scale-of-presence scores for System-1-like and System-2-like textual traces; $\text{RQS} \in [1,5]$ is a judge-scaled, text-grounded proxy for epistemic quality independent of conclusion polarity; the reasoning-balance index is $s_2 - s_1$; and the System-2 trace level is a $\{0,\ldots,4\}$ coarse marker of visible analytical elaboration.

{\sloppy
\textbf{Structured output.} The judge emits: the label; the three scalars; dimension scores; a primary rationale; three \texttt{supporting\_spans} with signal direction and a \textit{why}; up to three \texttt{counterevidence} entries; bias diagnostics (open-set, drawn from the eight-bias menu); a \texttt{nearest\_alternative\_label} plus \texttt{why\_not\_nearest\_alternative}; a composite $[0,1]$ score; and three derived indices (heuristic dominance, analytic strength, mixedness). The full system prompt and implementation are available in the \href{https://huggingface.co/spaces/nuojohnchen/Kahneman4Review}{demo code}.
\par}

\section{Additional analyses}
\label{app:extra}

\begin{figure}[!ht]
\centering
\includegraphics[width=0.65\linewidth]{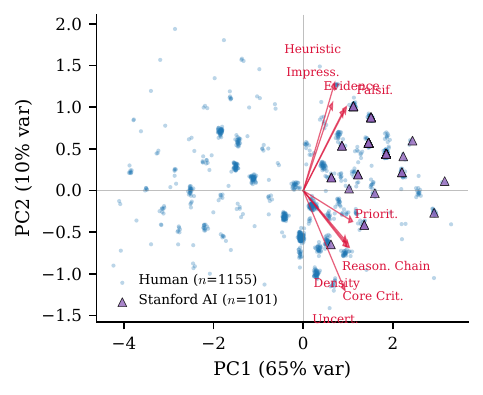}
\caption{PCA of the 9 rubric dimensions (S1-valenced dimensions sign-flipped). Fitted on human reviews; Stanford agentic reviews projected. Humans form a broad cluster around the origin; the agentic reviewer is shifted $+1.7$ on PC1 (the single analytical-trace axis). Arrows are dimension loadings.}
\label{fig:pca}
\end{figure}

\begin{figure}[!ht]
\centering
\includegraphics[width=0.65\linewidth]{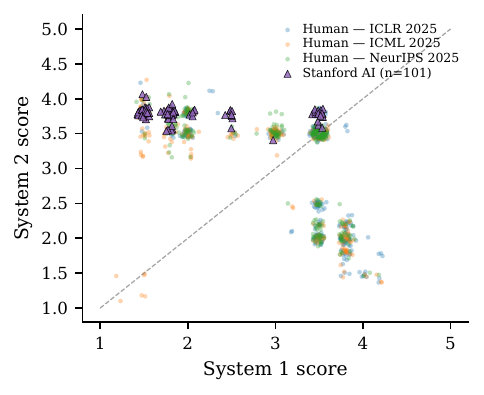}
\caption{Latent $(s_1, s_2)$ space. Human reviews cluster on the diagonal; the agentic AI reviewer (\textcolor{violet}{$\blacktriangle$}) is displaced up-and-left: simultaneously lower System-1 and higher System-2 traces.}
\label{fig:s1s2}
\end{figure}

\begin{figure}[!ht]
\centering
\includegraphics[width=0.7\linewidth]{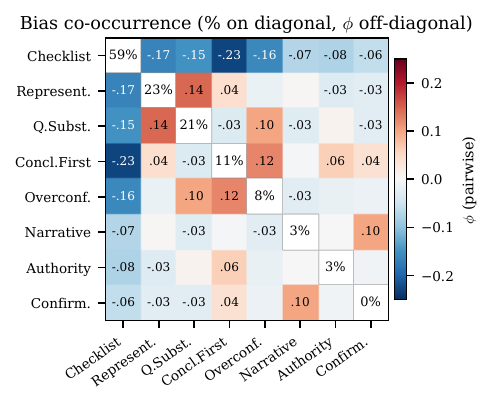}
\caption{Bias co-occurrence on human reviews. Diagonal is marginal frequency; off-diagonal is pairwise $\phi$ correlation. Positive clusters (Representativeness $\leftrightarrow$ Question Substitution, Conclusion-First $\leftrightarrow$ Overconfidence) support a weak co-occurring System-1-like trace pattern; most $|\phi|$ values are below 0.15.}
\label{fig:bias-cooccur}
\end{figure}

\section{Measurement validation details}
\label{app:validation}

We re-rate a fixed 100-review subset (deterministic seed 42, balanced by venue and label) under three conditions: two independent runs of \texttt{claude-3-7-sonnet-20250219} with a frozen system prompt and concurrency 4, and one run of \texttt{gpt-4o} with the same prompt. After dropping the small number of malformed-JSON outputs, $n{=}\ValN$ reviews are paired across all three; the resulting agreement statistics are reported in \cref{tab:validation}.

Two takeaways. (i) The earlier $\kappa{=}0.09$ stability figure was an artefact of prompt-version drift, not the rubric's actual stability under controlled re-runs. (ii) Cross-\emph{version} agreement is the weakest link: with the same frozen prompt, \texttt{sonnet-4-6} and \texttt{sonnet-3.7} produce materially different label distributions (notably, \texttt{sonnet-3.7} more frequently emits \textsc{Non-evaluative} for short reviews). This is a real model-version sensitivity, not a deprecation artefact: both models remain available and were used in this paper. The implication is that some headline numbers in \cref{sec:ai-vs-human} are model-version-specific, and reliability claims about LLM judges should be reported as model-conditional throughout.

\section{Length-controlled regression details}
\label{app:length}

The length-controlled model of \cref{eq:lengthreg} is fit by OLS with Stanford as the venue reference. Estimates: $\hat\beta_1 = \AICoefAdj$ (95\% CI $[\AICoefAdjLo, \AICoefAdjHi]$, $p{=}\AICoefAdjP$); $\hat\beta_2 = \LengthCoef$ per log-character. The naive RQS gap (AI $-$ human) is $\NaiveAIGap{}$; the $\hat\beta_1$ residual after adjustment is roughly $\PctGapExpl\%$ smaller. Conclusion: length is the largest single explainer of the headline effect, and the residual signal is real but small.

\section{Qualitative case studies}
\label{app:cases}

The following two rubric outputs are illustrative, not representative; we chose them because they exemplify the shape of a \textsc{System 1} human rating and a \textsc{System 2} agentic-reviewer rating. Raw review text is not reproduced (see \cref{app:data}).

\textbf{Case 1: Human, ICLR 2025 (\textsc{System 1}, RQS 1.8).}
\emph{Primary rationale.} ``The strengths section compresses the entire evaluation into two abstract labels (`novel' and `rigorous') without any supporting reasoning or evidence. The weaknesses are peripheral completeness requests (missing conclusion section, statistical tests) that are entirely disconnected from the paper's core claims about constrained optimization and weak supervision. There is no analytical linkage between any concern raised and the validity of the paper's central methodology.''
\emph{Bias diagnostics:} \texttt{Question Substitution}, \texttt{Checklist Inflation}.
\emph{Nearest alternative:} \textsc{Mixed} (rejected because the statistical-testing suggestion is the only partially analytical point, and it is brief and underdeveloped).

\textbf{Case 2: Stanford agentic reviewer, RQS 3.7.}
\emph{Primary rationale.} ``The passage is dominated by explicit, operational critique grounded in methodological reasoning. The weaknesses section in particular articulates specific technical concerns (Bayesian credible intervals used as frequentist confidence bounds under adaptive sampling, lack of union-over-time error control, absence of formal sample-complexity bounds), each with a clear causal link to why the gap undermines the paper's central claims.''
\emph{Bias diagnostics:} none.
\emph{Nearest alternative:} \textsc{Mixed}.

The pair supports the paper's central point: the agentic output exhibits the textual \emph{form} of methodological critique at a level the human rating does not. Whether the specific claims in Case 2 are \emph{correct} (whether adaptive-sampling intervals are really what the paper used, whether the cited concerns actually obtain) cannot be read off the rubric, which is why dimension-level scores on their own are insufficient for a trustworthiness benchmark.

\section{Short and template-inflated reviews}
\label{app:shortreviews}

A useful sanity check on what the rubric is responding to: do reviews that humans would intuitively recognise as textually underdeveloped map to \textsc{System 1}? We find a partial yes.

\textbf{Ultra-short reviews} (bottom 10\% of our sample by length, $\le \ShortThresh{}$ characters; $n=\NShortReviews$): mean RQS $\RQSShort$ vs $\RQSLong$ for the rest; $\PctShortSone\%$ are labelled \textsc{System 1} vs $\PctLongSone\%$. The dominant bias is \emph{Checklist Inflation} ($\PctShortTopBias\%$ of ultra-short reviews carry the tag), followed by \emph{Question Substitution}, the same pair that drives the broader co-occurring System-1-like trace pattern (\cref{fig:bias-cooccur}).

\textbf{Template-inflated long reviews.} A different textual failure mode is the ``40 numbered weaknesses, 40 numbered questions'' template, in which the surface specificity is high but the items are uniform peripheral asks (``please report standard deviations,'' ``please add multilingual evaluation,'' ``please compare with method X''). Two anonymised excerpts illustrate the pattern. Both reviewers gave the paper a low rating with maximum confidence.

\textit{Anonymised excerpt 1} (single-paragraph summary, then 40 numbered weaknesses, 40 numbered questions, near 1-to-1 mirrored):
\textit{``(31) The paper does not report the standard deviation of experimental results. (32) The paper's handling of background categories is unclear. \ldots (38) The paper's geometric feature calculation is not detailed. (39) The paper does not compare with multi-scale feature fusion methods. (40) The paper's conclusion overstates its contributions.''}

\textit{Anonymised excerpt 2} (40 numbered weaknesses + 40 mirrored questions on a different paper):
\textit{``(15) Inference speed on different hardware platforms is not provided. \ldots (28) Latency introduced by visual memory replay is not discussed. \ldots (40) Performance in multilingual scenarios is not tested.''}

Both reviewers issued ratings of ``4 (marginally below acceptance) / confidence 5 (absolutely certain),'' a confidence-rating combination that is itself a System-1-like textual signature (\emph{Overconfidence} co-occurring with \emph{Conclusion-First Justification}). The rubric labels both as \textsc{System 1} and tags both with \textit{Checklist Inflation} and \textit{Question Substitution}. A scan over all ICLR-2025 reviews in the bundle finds only nine with $\ge$30 numbered items in this template-inflated style; this is rare in absolute terms but is exactly the failure mode the rubric is designed to surface.

\textbf{Caveat: brevity is not low quality.} Ultra-short reviews are not labelled \textsc{System 1} mechanically: $16\%$ carry no bias tag in our sample. Some short reviews are terse precisely because the paper is straightforward and a long analysis is unwarranted. The rubric responds to the \emph{shape} of the analytic move, not to length per se; the length-controlled regression of \cref{app:length} confirms that, even within length strata, the signal that distinguishes the agentic AI reviewer from humans persists at smaller magnitude.

\section{Data, code, and reproducibility}
\label{app:data}

\textbf{Human reviews.} ICLR/ICML/NeurIPS 2025 reviews are drawn from publicly-available OpenReview decision pages; rating files reference reviews by OpenReview paper ID and reviewer ID. We do not redistribute review text verbatim due to OpenReview licence constraints. The longitudinal subset (ICLR 2021/2022/2023) follows the same convention.

\textbf{AI reviews.} The 101 agentic-reviewer outputs are drawn from a public showcase of an open agentic-reviewer system \cite{fars2026}, downloaded as a 102-file plaintext archive (FA0001--FA0234, with gaps); source details are available in the \href{https://huggingface.co/spaces/nuojohnchen/Kahneman4Review}{demo code}. The reviews are produced by submitting recent ML papers to the system; the corresponding source papers are not identified in the archive, so we use the reviews as a population-level distribution rather than as paper-paired comparisons to our human sample. We treat the resulting AI-vs-human gap as an upper bound on agentic-reviewer analytical-trace prevalence: if the showcase is curated for ``interesting'' or ``high-quality'' reviews, our sample over-represents the system's better outputs.

{\sloppy
\textbf{Code.} The released rubric prompt, implementation, ratings, and provenance information are available in the \href{https://huggingface.co/spaces/nuojohnchen/Kahneman4Review}{Kahneman4Review demo repository}.
\par}

\end{document}